# Island Loss for Learning Discriminative Features in Facial Expression Recognition

Jie Cai* Zibo Meng* Ahmed Shehab Khan Zhiyuan Li James O'Reilly Yan Tong
Department of Computer Science and Engineering, University of South Carolina, USA

*Abstract*— Over the past few years, Convolutional Neural Networks (CNNs) have shown promise on facial expression recognition. However, the performance degrades dramatically under real-world settings due to variations introduced by subtle facial appearance changes, head pose variations, illumination changes, and occlusions.

In this paper, a novel island loss is proposed to enhance the discriminative power of the deeply learned features. Specifically, the IL is designed to reduce the intra-class variations while enlarging the inter-class differences simultaneously. Experimental results on four benchmark expression databases have demonstrated that the CNN with the proposed island loss (IL-CNN) outperforms the baseline CNN models with either traditional softmax loss or the center loss and achieves comparable or better performance compared with the state-of-the-art methods for facial expression recognition.

## I. INTRODUCTION

As one of the most expressive parts of human, the face has been extensively studied in various active research fields. Automatic recognition of facial expression has attracted significant attention because of its importance and wide range of applications in human-computer interaction (HCI), such as interactive games, intelligent transportation, and animation. However, facial expression recognition in the wild is still a challenging problem because of high intra-class variations and high inter-class similarities caused by diversity in head pose, illumination, occlusions, and personal attributes.

As one of the major steps of facial expression recognition, features are extracted from either static images or videos to capture the appearance/geometry changes related to a target facial expression. In the past decades, various human crafted features have been adopted for facial expression recognition. Most recently, deep features learned by deep convolutional neural networks (CNNs) have achieved promising results on facial expression recognition especially in more challenging settings [16], [48], [30], [8], [46], [3].

Traditional CNNs are optimized using a softmax loss, which penalizes the misclassified samples and thus forces the features of different classes staying apart. As illustrated in Fig. 1 (a), the learned features form clusters corresponding to different expressions in the feature space. However, due to high intra-class variations, the features in each cluster are often scattered. Furthermore, the clusters overlap because of high inter-class similarities. Most recently, an additional center loss was introduced into CNNs [44] to reduce the intra-class variations of the learned features for face recognition. As shown in Fig. 1(b), the samples are pulled towards

* indicates equal contribution.

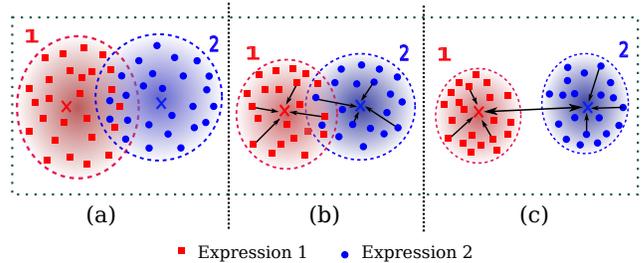

Fig. 1. An illustration of deep features learned by (a) a softmax loss, (b) a softmax loss + a center loss, and (c) a softmax loss + an island loss in the feature space. Inclusion of a center loss pulls the features of the same expression towards their centers denoted by a cross; while the island loss not only compresses the clusters individually, but also pushes the two clusters apart. Best viewed in color.

their corresponding centers with smaller intra-class variations compared to those learned only using softmax loss. However, the inter-class similarity was not considered in the center loss. This motivates us to further enhance the discriminative power of the learned deep features by increasing the differences between different expressions. Specifically, as depicted in Fig. 1(c), we propose an island loss to simultaneously compress each cluster and push cluster centers apart as isolated "islands".

To demonstrate the effectiveness of the proposed island loss, a CNN with the island loss (IL-CNN) is developed for facial expression recognition. As illustrated in Fig. 2, the IL-CNN architecture includes three convolutional layers, each of which is followed by a PReLU layer and a batch normalization (BN) layer. A max pooling layer is employed after each of the first two BN layers. Following the third convolutional layer, two fully-connected layers are used to generate the representation for each input sample. An island loss is calculated at the second fully-connected layer. Finally, a softmax loss is calculated at the decision layer to produce the distribution over the target expressions and to calculate the classification errors. The island loss and the softmax loss are jointly minimized to drive the fine-tuning process in the CNN training.

In summary, our major contributions are:
- Proposing a novel loss function with the island loss, which aims to learn representations with lower intra-class variations and higher inter-class distances; and
- Developing an IL-CNN with the proposed island loss to learn discriminative features for facial expression

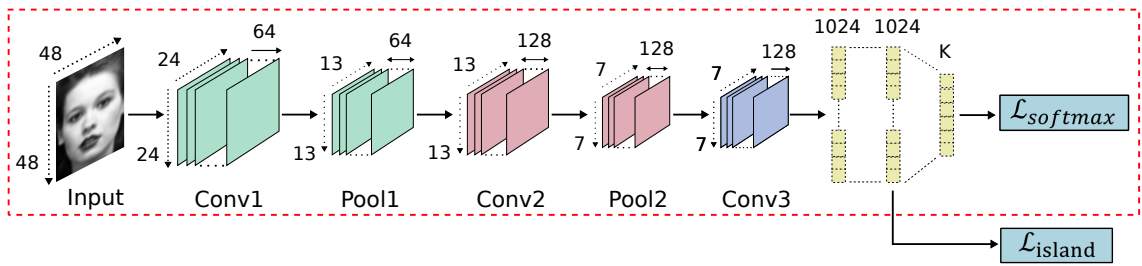

Fig. 2. The proposed IL-CNN for facial expression recognition. An island loss calculated at the second fully-connected layer and the softmax loss calculated at the decision layer are responsible to fine-tuning the CNN parameters. Best viewed in color.

recognition.

The proposed IL-CNN was evaluated on three well-known posed facial expression databases, i.e. Extended Cohn-Kanade database (CK+) [14], [24], Oulu-CASIA database [50] and MMI database [31]. More importantly, it was also evaluated on a spontaneous facial expression dataset, i.e. Static Facial Expressions in the Wild (SFEW) [6], which contains face images with large head pose variations and different illuminations and is widely used for benchmarking facial expression recognition in the wild. Experimental results on these four databases have shown that the proposed IL-CNN outperforms the baseline CNN models using the traditional softmax loss or the center loss, thanks to the increased inter-class distances and further reduced intra-class variations compared to that using the center loss. It also achieves comparable or better performance compared to the state-of-the-art facial expression recognition methods.

## II. RELATED WORK

Facial activity analysis has been widely studied as detailed in the recent surveys [35], [25]. One of the major steps is to extract the most discriminative features that characterize facial appearance and geometry changes caused by facial behavior. These features can be roughly divided into two categories: human designed and learned features.

Gabor wavelets [4], Scale Invariant Feature Transform (SIFT) features [49], histogram of Oriented Gradients (HOG) [2], histograms of Local Binary Patterns (LBP) [41], [15], histograms of Local Phase Quantization (LPQ) [11], histograms of Local Gabor Binary Patterns (LGBP) [29] have been demonstrated to be the most successful human designed features. In addition to the spatiotemporal extensions of the aforementioned 2D features [17], [37], [51], [12], [45], [50], features are deliberately designed to utilize both spatial and temporal information in an image sequence, such as temporal modeling of shapes (TMS) [10], interval temporal Bayesian network [43], expressionlets on spatiotemporal manifold (STM-ExpLet) [21], selective transfer machine (STM) [5], Gabor phase shifts (F-Bases) [36], Latent Ordinal Model (LOMo) [38], spatiotemporal covariance descriptors (Cov3D) [34].

Benefiting from the advance in feature learning, features can be learned either unsupervised by sparse coding [23], [53], [32], [27] or supervised by deep learning [33], [40], [22], [7], [28], [19], [13], [20], [52], [8], [46], [3]. Among them, the deep CNNs have achieved promising recognition performance under real-world conditions as demonstrated in the recent EmotiW2015 [16], [48], [30], [47], [39], [54] and EmotiW2016 challenge [8], [46], [3].

Most of the aforementioned CNN-based approaches adopted the softmax loss as the supervision signal to train the CNN models. In real-world scenarios, facial expression recognition suffers from high intra-class variations and inter-class similarities. On one hand, CNNs may generate similar representations for images containing the different expressions, especially for the same person. On the other hand, the deeply learned features may be different for images containing the same expression resulted from the aforementioned challenges. Recently, an Identity-Aware CNN (IACNN) [26] was developed to alleviate variations introduced by personal attributes using an expression-sensitive contrastive loss and an identity-sensitive contrastive loss. However, the variations caused by other factors such as head pose and illumination were not considered. In addition, the contrastive loss suffers from drastic data expansion when constructing image pairs from the training set. Wen et al. [44] introduced a center loss for face recognition, which targets directly on one of the learning objectives, i.e., the intra-class compactness. As a result, the learned features from the same subject will be more similar. Rather than minimizing the distance to the cluster center as the center loss, Li et al. [18] proposed a Deep Locality-Preserving CNN (DLP-CNN), which preserves the locality proximity by minimizing the distance to the K-nearest neighbors within the same class. While the center loss pulls the samples towards their corresponding cluster centers and DLP-CNN pushes the samples to their K-nearest neighbors, the proposed island loss further enhances the discriminative power of the deep features by simultaneously reducing intra-class variations and augmenting inter-class differences as demonstrated in the experiments on four expression databases.

## III. METHODOLOGY

In this section, we will first give a brief review of the center loss, and then introduce the proposed island loss followed by the forward and backward propagation processes of the IL-CNN.

## A. A Brief Review of Center Loss

As illustrated in Fig. 1(b), the center loss [44] explicitly reduces the intra-class variations by pushing samples towards their corresponding class centers in the feature space during training. The centers will be updated in each iteration using Stochastic Gradient Descent (SGD) as part of the CNN training.

*1) Forward propagation:* The center loss denoted as $\mathcal{L}_C$ is defined in Eq. 1 [44] as the summation of squared distances between samples and their corresponding centers in the feature space:

$$\mathcal{L}_C = \frac{1}{2} \sum_{i=1}^{m} ||\mathbf{x}_i - \mathbf{c}_{y_i}||^2 \quad (1)$$

where $y_i$ is the class label of the $i^{th}$ sample; $\mathbf{x}_i$ denotes the feature vector of the $i^{th}$ sample taken from the fully-connected layer before the decision layer; $\mathbf{c}_{y_i} \in \mathcal{R}^d$ denotes the center of all samples with the same class label as $y_i$; and $m$ is the number of samples in the mini-batch. By minimizing the center loss, the samples of the same class will be pulled towards their corresponding centers and thus, the overall intra-class variations can be reduced.

During forward propagation, a joint loss is calculated as the weighted sum of the softmax loss and the center loss, which is used in the backward propagation to drive the fine-tuning process:

$$\mathcal{L} = \mathcal{L}_S + \lambda \mathcal{L}_C \quad (2)$$

where $\mathcal{L}_S$ is the softmax loss; and a scalar $\lambda$ is used for balancing the softmax loss and the center loss.

*2) Backward propagation:* During backward propagation, the partial derivative of the center loss $\mathcal{L}_C$ with respect to the input sample $\mathbf{x}_i$ can be calculated as

$$\frac{\partial \mathcal{L}_c}{\partial \mathbf{x}_i} = \mathbf{x}_i - \mathbf{c}_{y_i} \quad (3)$$

In addition, the centers will be updated in the iterative optimization of the CNN using SGD as

$$\Delta \mathbf{c}_j = \frac{\sum_{i=1}^{m} \delta(y_i, j)(\mathbf{c}_j - \mathbf{x}_i)}{1 + \sum_{i=1}^{m} \delta(y_i, j)} \quad (4)$$

where $\delta(y_i, j)$ is defined as

$$\delta(y_i, j) = \begin{cases} 1, & y_i = j \\ 0, & y_i \neq j \end{cases} \quad (5)$$

## B. An Island Loss for Facial Expression Recognition

As shown in Fig. 1(b), minimizing the center loss tends to reduce the intra-class variations of the deep features, while the clusters from different classes may be overlapped with each other. To cope with this problem, an island loss is proposed to reduce the intra-class variations and increase the inter-class differences simultaneously.

*1) Forward propagation:* The island loss denoted as $\mathcal{L}_{IL}$ is defined as the summation of the center loss and the pairwise distances between class centers in the feature space:

$$\mathcal{L}_{IL} = \mathcal{L}_C + \lambda_1 \sum_{\mathbf{c}_j \in \mathcal{N}} \sum_{\substack{\mathbf{c}_k \in \mathcal{N} \\ \mathbf{c}_k \neq \mathbf{c}_j}} \left( \frac{\mathbf{c}_k \cdot \mathbf{c}_j}{||\mathbf{c}_k||_2 ||\mathbf{c}_j||_2} + 1 \right) \quad (6)$$

where $\mathcal{N}$ is the set of expression labels; $\mathbf{c}_k$ and $\mathbf{c}_j$ denote the $k^{th}$ and $j^{th}$ center with $L_2$ norm $||\mathbf{c}_k||_2$ and $||\mathbf{c}_j||_2$, respectively; $(\cdot)$ represents the dot product. Specifically, the first term penalizes the distance between the sample and its corresponding center and the second term penalizes the similarity between expressions. $\lambda_1$ is used for balancing the two terms. By minimizing the island loss, the samples of the same expression will get closer to each other and those of different expressions will be pushed apart.

The overall loss function of CNN training is given by Eq. 7:

$$\mathcal{L} = \mathcal{L}_S + \lambda \mathcal{L}_{IL} \quad (7)$$

where a hyper parameter $\lambda$ is used to balance the two losses.

*2) Backward propagation:* The partial derivative of the island loss $\mathcal{L}_{IL}$ with respect to the input sample $x_i$ can be calculated as

$$\frac{\partial \mathcal{L}_{IL}}{\partial \mathbf{x}_i} = \mathbf{x}_i - \mathbf{c}_{y_i} \quad (8)$$

which is actually the same as the one only using the center loss as in Eq. 3. $\frac{\partial \mathcal{L}_{IL}}{\partial \mathbf{x}_i}$ will be further backpropagated to the lower fully-connected layer and the convolutional layers to drive the fine-tuning process of CNNs.

**Updating the cluster center:** Based on SGD, the update of the $j^{th}$ class center can be calculated as

$$\Delta \mathbf{c}_j = \frac{\sum_{i=1}^{m} \delta(y_i, j)(\mathbf{c}_j - \mathbf{x}_i)}{1 + \sum_{i=1}^{m} \delta(y_i, j)} + \frac{\lambda_1}{|\mathcal{N}| - 1} \sum_{\substack{\mathbf{c}_k \in \mathcal{N} \\ \mathbf{c}_k \neq \mathbf{c}_j}} \frac{\mathbf{c}_k}{||\mathbf{c}_k||_2 ||\mathbf{c}_j||_2} - \left( \frac{\mathbf{c}_k \cdot \mathbf{c}_j}{||\mathbf{c}_k||_2 ||\mathbf{c}_j||_2^3} \right) \mathbf{c}_j \quad (9)$$

where $|\mathcal{N}|$ is the total number of expressions.

In this manner, the class centers can be updated iteratively in each mini-batch with a learning rate $\alpha$ [1]:

$$\mathbf{c}_j^{t+1} = \mathbf{c}_j^t - \alpha \Delta \mathbf{c}_j^t \quad (10)$$

The forward and backward learning process in the IL-CNN is summarized in Algorithm 1.

## IV. EXPERIMENTS

A series of experiments have been conducted on four benchmark expression databases including three posed facial expression databases, i.e. the CK+ database [14], [24], the MMI database [31], and the Oulu-CASIA database [50], and more importantly, on a spontaneous facial expression database, i.e., the SFEW dataset [6] to evaluate the proposed IL-CNN for facial expression recognition. Furthermore, to demonstrate the effectiveness of the proposed island loss, the IL-CNN is compared with two baseline CNNs, which have the same network structure as the IL-CNN, but are under the supervision of (1) softmax loss and (2) softmax loss + center loss, respectively.

---
[1] In our experiments, we set $\alpha = 1$, $\lambda = 0.01$, $\lambda_1 = 10$ empirically.

**Algorithm 1** Forward-backward learning algorithm of IL-CNN
**Input:** Training data $\{\mathbf{x}_i\}$.
1: **Given:** mini-batch size $m$, number of iterations $T$, learning rates $\mu$ and $\alpha$, and hyper-parameters $\lambda$ and $\lambda_1$.
2: **Initialize:** $t = 1$, network layer parameters $\mathcal{W}$, softmax loss parameters $\theta$, and island loss parameters $\mathbf{c}_j$.
3: **for** $t = 1$ to $T$ **do**
4:    Calculate the joint loss as in Eq. 7:
5:      $\mathcal{L} = \mathcal{L}_S + \lambda \mathcal{L}_{IL}$
6:    Update the softmax loss parameters:
7:      $\theta^{t+1} = \theta^t - \mu \frac{\partial \mathcal{L}_S^t}{\partial \theta^t}$
8:    Update the island loss parameters (i.e. centers) as in Eq. 10:
9:      $\mathbf{c}_j^{t+1} = \mathbf{c}_j^t - \alpha \Delta \mathbf{c}_j^t$
10:   Update the backpropagation error:
11:     $\frac{\partial \mathcal{L}^t}{\partial \mathbf{x}_i^t} = \frac{\partial \mathcal{L}_S^t}{\partial \mathbf{x}_i^t} + \lambda \frac{\partial \mathcal{L}_{IL}^t}{\partial \mathbf{x}_i^t}$
12:   Update the network layer parameters:
13:     $\mathcal{W}^{t+1} = \mathcal{W}^t - \mu \frac{\partial \mathcal{L}^t}{\partial \mathcal{W}^t} = \mathcal{W}^t - \mu \frac{\partial \mathcal{L}^t}{\partial \mathbf{x}_i^t} \frac{\partial \mathbf{x}_i^t}{\partial \mathcal{W}^t}$
14: **end for**
**Output:** Network layer parameters $\mathcal{W}$, island loss parameters $\mathbf{c}_j$, and softmax loss parameters $\theta$.

### A. Preprocessing

To reduce the variations in face scale and in-plane rotation, face alignment is employed on each image based on the facial landmarks extracted by Discriminative Response Map Fitting (DRMF) [1]. Specifically, given the 66 extracted facial landmarks, face regions are aligned based on three key points: centers of two eyes and mouth. The aligned facial images are then resized to $60 \times 60$. In addition, histogram equalization is utilized to improve the contrast in facial images. Because of the limited number of images in the facial expression databases, a data augmentation strategy is adopted to produce more data for training. Specifically, $48 \times 48$ patches are randomly cropped from the $60 \times 60$ images, and then rotated by a random degree between -10° and 10°. The rotated images are randomly horizontally flipped as the input of all CNNs, resulting in a new dataset 5,760 times larger than the original one.

### B. Experimental Datasets

*1) CK+ Dataset:* The CK+ database [14], [24] consists of 327 videos collected from 118 subjects, each of which is associated with one of 7 expression labels, i.e. anger, contempt, disgust, fear, happiness, sadness, and surprise. Each video starts with a neutral face, and reaches the peak in the last frame. To collect more data, the last three frames of each sequence are collected associated with the provided expression label. Thus, an experimental dataset consisting of 981 images is built.

*2) MMI dataset:* The MMI database [31] contains 213 image sequences, from which 208 sequences with frontal-view faces of 31 subjects are used in our experiment. Each sequence is labeled as one of six basic expressions, i.e. anger, disgust, fear, happiness, sadness, and surprise, starting from a neutral expression, through a peak phase in the middle, and back to a neutral face at the end. Since the actual location of the peak frame is not provided, three frames in the middle of each image sequence are collected as peak frames associated with the provided label. Hence, there are a total of 624 images used in our experiments.

*3) Oulu-CASIA dataset:* The Oulu-CASIA database [50] contains 2,880 videos, each of which contains one of the six basic expressions (happiness, sadness, surprise, anger, fear, and disgust), collected from 80 subjects. Each of the videos is captured with one of two imaging systems, i.e. near infrared (NIR) and visible light (VIS), under one of three different illumination conditions: normal indoor illumination, weak illumination or dim illumination. Following the previous work evaluated on the Oulu-CASIA database [13], only the 480 videos collected by the VIS System under normal indoor illumination are employed in our experiments. For each video, the last three frames are collected as the peak frames of the labeled expression. Thus, the Oulu-CASIA dataset contains 1,440 images for our experiments.

**Training/testing strategy:** The baseline CNNs and the proposed IL-CNN are trained and tested on static images. A 10-fold cross-validation strategy is employed for CK+, MMI, and Oulu-CASIA datasets, where each dataset is further split into 10 subsets, and the subjects in any two subsets are mutually exclusive. For each run, data from 8 sets are used for training, the remaining two subsets are used for validation and testing, respectively. The final sequence-level decision is made by choosing the label of the class with the highest average score over the three images from the same sequence. The results are reported as the average of the 10 runs.

*4) SFEW dataset:* The SFEW database [6] is the most widely used benchmark database for facial expression recognition in the wild. It is composed of 1,766 images, i.e. 958 for training, 436 for validation, and 372 for testing. Each of the images has been assigned to one of seven expression categories, i.e., anger, disgust, fear, neutral, happy, sad, and surprise. The expression labels of the training and validation sets are provided, while those of the testing set is held back by the challenge organizer. Thus, the performance on the testing set is evaluated and provided by the challenge organizer.

### C. CNN Implementation Details

For experiments on each benchmark dataset, a CNN with the softmax loss is pre-trained using the Facial Expression Recognition (FER-2013) dataset [9] and the other three datasets. Starting from the same pre-trained CNN, the IL-CNN and the two baseline CNNs are fine-tuned on each

TABLE I

CONFUSION MATRIX OF THE PROPOSED IL-CNN EVALUATED ON THE CK+ DATASET [14], [24]. THE GROUND TRUTH AND THE PREDICTED LABELS ARE GIVEN BY THE FIRST COLUMN AND THE FIRST ROW, RESPECTIVELY.

|    | An    | Co    | Di    | Fe    | Ha    | Sa    | Su    |
|----|-------|-------|-------|-------|-------|-------|-------|
| An | **95.6%** | 2.2%  | 0%    | 0%    | 0%    | 2.2%  | 0%    |
| Co | 11.1% | **74.1%** | 0%    | 3.7%  | 3.7%  | 7.4%  | 0%    |
| Di | 0%    | 0%    | **100%** | 0%    | 0%    | 0%    | 0%    |
| Fe | 0%    | 4%    | 0%    | **84.0%** | 12.0% | 0%    | 0%    |
| Ha | 0%    | 0%    | 0%    | 0%    | **100%** | 0%    | 0%    |
| Sa | 14.3% | 0%    | 3.6%  | 0%    | 0%    | **82.1%** | 0%    |
| Su | 0%    | 1.2%  | 0%    | 0%    | 0%    | 0%    | **98.8%** |

of the four benchmark expression datasets, respectively. Stochastic gradient descent with a momentum of 0.9, a mini-batch size of 300, and a weight decay parameter of 0.05, is used for training the CNNs. The learning rate $\mu$ starts from 0.001 and is reduced by a factor of 0.1 for every 2500 iterations. A dropout rate of 0.6 is employed for the last two fully-connected layers, i.e. zeroing out the output of a neuron with a probability of 0.6.

*D. Experimental Results*

The confusion matrices of the proposed IL-CNN are reported for the four datasets, where entries along the diagonal represent the recognition accuracy for each expression. In addition to the two baseline CNNs, the proposed IL-CNN is compared with the state-of-the-art methods evaluated on the four databases such as methods using human crafted features (HOG 3D [17], TMS [10], Cov3D [34], STM [5], STM-ExpLet [21], LOMo [38], ITBN [43] and F-Bases [36]), a method using sparse coding (MSR [32]), and CNN-based methods (3DCNN and 3DCNN-DAP [20], Inception [28], IACNN [26], DLP-CNN [18], FN2EN [7], PPDN [52] and DTAGN [13]).

*1) Results on CK+ dataset:* The confusion matrix of the proposed IL-CNN is reported in Table I and the comparison results in terms of average recognition accuracy are reported in Table II. The IL-CNN achieves an average recognition accuracy of 94.35% on the CK+ dataset for the 7 expressions.

*2) Results on MMI dataset:* The confusion matrix of the proposed IL-CNN is given by Table III and the comparison results are presented in Table IV for the MMI dataset.

*3) Results on Oulu-CASIA dataset:* Table V gives the confusion matrix of the IL-CNN and Table VI summarizes the comparison results on the Oulu-CASIA dataset.

**Result analysis on the three posed facial expression datasets:** As shown in Table II, Table IV, and Table VI, the IL-CNN consistently outperforms the two baseline CNNs using either softmax loss or center loss by reducing the intra-class variations and inter-class similarities. Furthermore, the IL-CNN also achieves better or at least comparable performance compared to the state-of-the-art methods. Note that, while most of the state-of-the-art methods utilized dynamic features extracted from image sequences, the proposed IL-CNN is trained on static images, which is more favorable

TABLE II

PERFORMANCE COMPARISON ON THE CK+ DATASET [14], [24] IN TERMS OF THE AVERAGE RECOGNITION ACCURACY OF 7 EXPRESSIONS.

| Method | Acc. | Classes | Feature | Strategy |
|---|---|---|---|---|
| 3DCNN [20] | 85.9 | 7 | Dynamic | 15 folds |
| ITBN [43] | 86.3 | 7 | Dynamic | 15 folds |
| F-Bases [36] | 89.01 | 7 | Dynamic | LOSO |
| MSR [32] | 91.4 | 7 | Static | LOSO |
| HOG 3D [17] | 91.44 | 7 | Dynamic | 10 folds |
| TMS [10] | 91.89 | 6 | Dynamic | 4 folds |
| Cov3D [34] | 92.3 | 7 | Dynamic | 5 folds |
| 3DCNN-DAP [20] | 92.4 | 7 | Dynamic | 15 folds |
| Inception [28] | 93.2 | 6 | Static | 5 folds |
| STM-ExpLet [21] | 94.19 | 7 | Dynamic | 10 folds |
| F-Bases [36] | 94.81 | 7 | Static | LOSO |
| LOMo [38] | 95.1 | 7 | Dynamic | 10 folds |
| IACNN [26] | 95.37 | 7 | Static | 8 folds |
| DLP-CNN [18] | 95.78 | 6 | Static | 5 folds |
| STM [5] | 96.40 | 7 | Dynamic | N/A |
| FN2EN [7] | 96.80 | 8 | Static | 10 folds |
| DTAGN [13] | 97.25 | 7 | Dynamic | 10 folds |
| PPDN [52] | **97.3** | 6 | Static | 10 folds |
| softmax loss | 91.07 | 7 | Static | 10 folds |
| center loss | 92.26 | 7 | Static | 10 folds |
| **IL-CNN** | **94.35** | 7 | Static | 10 folds |

TABLE III

CONFUSION MATRIX OF THE PROPOSED IL-CNN EVALUATED ON THE MMI DATASET [31]. THE GROUND TRUTH AND THE PREDICTED LABELS ARE GIVEN BY THE FIRST COLUMN AND THE FIRST ROW, RESPECTIVELY.

|    | An    | Di    | Fe    | Ha    | Sa    | Su    |
|----|-------|-------|-------|-------|-------|-------|
| An | **66.7%** | 12.1% | 6.1%  | 3.0%  | 12.1% | 0%    |
| Di | 9.4%  | **78.1%** | 6.3%  | 6.3%  | 0%    | 0%    |
| Fe | 10.7% | 7.1%  | **57.1%** | 7.1%  | 3.6%  | 14.3% |
| Ha | 0%    | 2.4%  | 4.8%  | **92.9%** | 0%    | 0%    |
| Sa | 18.8% | 6.3%  | 15.6% | 0%    | **59.4%** | 0%    |
| Su | 7.3%  | 0%    | 26.8% | 2.4%  | 0%    | **63.4%** |

TABLE IV

PERFORMANCE COMPARISON ON THE MMI DATASET [31] IN TERMS OF THE AVERAGE RECOGNITION ACCURACY OF 6 EXPRESSIONS.

| Method | Acc. | Classes | Feature | Strategy |
|---|---|---|---|---|
| 3DCNN [20] | 53.2 | 6 | Dynamic | 20 folds |
| ITBN [43] | 59.7 | 6 | Dynamic | 20 folds |
| HOG 3D [17] | 60.89 | 6 | Dynamic | 10 folds |
| 3DCNN-DAP [20] | 63.4 | 6 | Dynamic | 20 folds |
| DTAGN [13] | 70.24 | 6 | Dynamic | 10 folds |
| IACNN [26] | 71.55 | 6 | Static | 10 folds |
| STM-ExpLet [21] | 75.12 | 6 | Dynamic | 10 folds |
| F-Bases [36] | 57.56 | 6 | Static | LOSO |
| F-Bases [36] | 73.66 | 6 | Dynamic | LOSO |
| Inception [28] | **77.6** | 6 | Static | 5 folds |
| softmax loss | 66.35 | 6 | Static | 10 folds |
| center loss | 69.23 | 6 | Static | 10 folds |
| **IL-CNN** | **70.67** | 6 | Static | 10 folds |

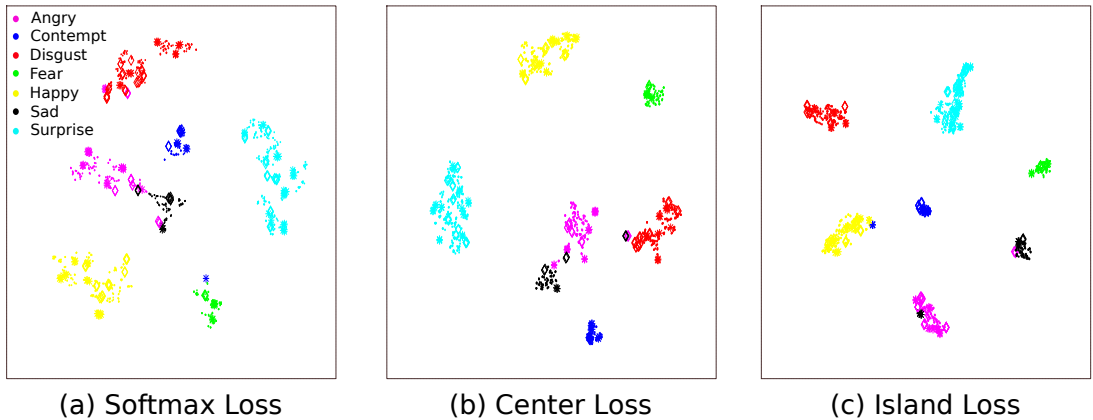

Fig. 3. A visualization study of the deep features learned by CNNs using (a) softmax loss, (b) softmax + center loss, and (c) softmax + island loss on the CK+ database. There are a total of 981 samples including 263 × 3 training data from 8 subsets, 32 × 3 validation data, and 32 × 3 testing data. The dots, stars, and diamonds represent training, validation, and testing data, respectively. Note that the features learned by the IL-CNN are well separated according to expressions. Best viewed in color.

TABLE V
CONFUSION MATRIX OF THE PROPOSED IL-CNN EVALUATED ON THE OULU-CASIA DATABASE [50]. THE GROUND TRUTH AND THE PREDICTED LABELS ARE GIVEN BY THE FIRST COLUMN AND THE FIRST ROW, RESPECTIVELY.

|    | An    | Di    | Fe    | Ha    | Sa    | Su    |
|----|-------|-------|-------|-------|-------|-------|
| An | **67.5%** | 12.5% | 5.0%  | 1.3%  | 13.8% | 0%    |
| Di | 22.5% | **70.0%** | 1.3%  | 0%    | 6.3%  | 0%    |
| Fe | 8.8%  | 2.5%  | **68.8%** | 11.3% | 2.5%  | 6.3%  |
| Ha | 0%    | 0%    | 6.3%  | **92.5%** | 0%    | 1.3%  |
| Sa | 13.8% | 2.5%  | 6.3%  | 1.3%  | **76.3%** | 0%    |
| Su | 3.8%  | 0%    | 7.5%  | 0%    | 0%    | **88.8%** |

TABLE VI
PERFORMANCE COMPARISON ON THE OULU-CASIA DATABASE [50] IN TERMS OF THE AVERAGE RECOGNITION ACCURACY OF 6 EXPRESSIONS.

| Method | Acc. | Classes | Feature | Strategy |
|--------|------|---------|---------|----------|
| HOG 3D [17] | 70.63 | 6 | Dynamic | 10 folds |
| AdaLBP [50] | 73.54 | 6 | Dynamic | 10 folds |
| STM-ExpLet [21] | 74.59 | 6 | Dynamic | 10 folds |
| DTAGN [13] | 81.46 | 6 | Dynamic | 10 folds |
| LOMo [38] | 82.1 | 6 | Dynamic | 10 folds |
| FN2EN [7] | **87.71** | 6 | Static | 10 folds |
| softmax loss | 73.54 | 6 | Static | 10 folds |
| center loss | 75.63 | 6 | Static | 10 folds |
| **IL-CNN** | 77.29 | 6 | Static | 10 folds |

TABLE VII
CONFUSION MATRIX OF THE PROPOSED IL-CNN EVALUATED ON THE SFEW [6] VALIDATION SET. THE GROUND TRUTH AND THE PREDICTED LABELS ARE GIVEN BY THE FIRST COLUMN AND THE FIRST ROW, RESPECTIVELY.

|    | An    | Di    | Fe    | Ha    | Ne    | Sa    | Su    |
|----|-------|-------|-------|-------|-------|-------|-------|
| An | **61.0%** | 0%    | 1.3%  | 1.3%  | 22.1% | 11.7% | 2.6%  |
| Di | 4.4%  | **0%**    | 0%    | 13.0% | 30.4% | 43.5% | 8.7%  |
| Fe | 6.4%  | 2.1%  | **6.4%**  | 6.4%  | 55.3% | 8.5%  | 14.9% |
| Ha | 0%    | 0%    | 1.4%  | **89.0%** | 5.5%  | 4.1%  | 0%    |
| Ne | 8.1%  | 0%    | 1.2%  | 1.2%  | **66.2%** | 22.1% | 1.2%  |
| Sa | 4.1%  | 0%    | 0%    | 0%    | 43.8% | **48.0%** | 4.1%  |
| Su | 14.0% | 0%    | 1.8%  | 5.3%  | 38.6% | 7.0%  | **33.3%** |

TABLE VIII
CONFUSION MATRIX OF THE PROPOSED IL-CNN EVALUATED ON THE SFEW [6] TESTING SET. THE GROUND TRUTH AND THE PREDICTED LABELS ARE GIVEN BY THE FIRST COLUMN AND THE FIRST ROW, RESPECTIVELY.

|    | An    | Di    | Fe    | Ha    | Ne    | Sa    | Su    |
|----|-------|-------|-------|-------|-------|-------|-------|
| An | **73.9%** | 0%    | 2.9%  | 0%    | 20.3% | 1.5%  | 1.5%  |
| Di | 17.7% | **11.8%** | 0%    | 17.7% | 11.8% | 41.2% | 0%    |
| Fe | 17.1% | 2.4%  | **17.1%** | 0%    | 46.3% | 12.2% | 4.9%  |
| Ha | 3.2%  | 2.1%  | 0%    | **73.7%** | 14.7% | 6.3%  | 0%    |
| Ne | 8.6%  | 0%    | 0%    | 0%    | **65.5%** | 24.1% | 1.7%  |
| Sa | 1.8%  | 3.6%  | 9.1%  | 0%    | 36.4% | **47.3%** | 1.8%  |
| Su | 5.4%  | 2.7%  | 8.1%  | 2.7%  | 27.0% | 5.4%  | **48.7%** |

for online applications or snapshots where per frame labels are preferred. We are aware that the Inception model [28] has the best performance on the MMI database owing to a much more complex and deeper network structure. The proposed island loss can be adopted by these advanced network structures by replacing the softmax loss.

*4) Results on SFEW dataset:* The confusion matrices of IL-CNN are given by Table VII and Table VIII for the validation and testing sets of SFEW, respectively. As illustrated in Table IX, the proposed IL-CNN outperforms the baseline CNNs for both validation set and the testing set. Furthermore, the IL-CNN, which uses a single CNN with a shallow architecture, is ranked at the third place for the testing set among all the methods compared. Note that, Kim et al. [16] and Yu et al. [48], who are ranked the $1^{st}$ and $2^{nd}$, utilized an ensemble of CNNs. In addition, Yu et al. [48] also employed a combination of different network structures. Thus, an ensemble of IL-CNNs has been constructed and achieves comparable performance as the best methods [16], [48] on the SFEW dataset.

TABLE IX
PERFORMANCE COMPARISON ON THE SFEW DATABASE [6] IN TERMS OF THE AVERAGE RECOGNITION ACCURACY OF 7 EXPRESSIONS.

| Method | Validation Set | Test Set |
|---|---|---|
| Kim et al. [16] | 53.9 | **61.6** |
| Yu et al. [48] | **55.96** | 61.29 |
| Ng et al. [30] | 48.5 | 55.6 |
| Yao et al. [47] | 43.58 | 55.38 |
| Sun et al. [39] | 51.02 | 51.08 |
| Zong et al. [54] | N/A | 50 |
| Kaya et al. [15] | 53.06 | 49.46 |
| Mollahosseini et al. [28] | 47.7 | N/A |
| Dhall et al. [6] (baseline of SFEW) | 35.93 | 39.13 |
| STTLDA [55] | N/A | 50.00 |
| IACNN [26] | 50.98 | 54.30 |
| DLP-CNN [18] | 51.05 | N/A |
| FN2EN [7] | 55.15 | N/A |
| softmax loss | 47.94 | 52.12 |
| center loss | 48.85 | 53.79 |
| IL-CNN | 51.83 | 56.98 |
| **ensemble IL-CNN** | **52.52** | **59.41** |

TABLE X
AVERAGE BETWEEN-CENTER-DISTANCE ON FOUR DATASETS. THE BIGGER VALUE MEANS THE FURTHER DISTANCE.

| Dataset | CK+ | MMI | CASIA | SFEW Val |
|---|---|---|---|---|
| Softmax loss | 0.3418 | 0.2934 | 0.3537 | 0.3012 |
| Center loss | 0.3665 | 0.3165 | 0.3668 | 0.3274 |
| Island loss | **0.4396** | **0.5250** | **0.4562** | **0.3778** |

TABLE XI
AVERAGE SAMPLE-TO-CENTER DISTANCE ON FOUR DATASETS. THE SMALLER VALUE MEANS THE CLOSER DISTANCE.

| Dataset | CK+ | MMI | CASIA | SFEW Val |
|---|---|---|---|---|
| Softmax loss | 0.0301 | 0.0804 | 0.0536 | 0.0797 |
| Center loss | 0.0233 | 0.0634 | 0.0395 | 0.0694 |
| Island loss | **0.0194** | **0.0618** | **0.0307** | **0.0689** |

### E. Visualization Study

A visualization study is performed to demonstrate that the island loss is effective in reducing intra-class variations while increasing the inter-class differences. Specifically, the features learned by the two baseline CNNs and the IL-CNN are visualized using t-SNE [42], which is widely employed to visualize high dimensional data. As illustrated in Fig. 3, the learned features are clustered according to 7 expressions, where the training, validation, and testing samples are denoted by dots, stars and diamonds, respectively.

By employing the center loss, samples of the same expression are closer to each other in Fig. 3b compared with those learned with only the softmax loss (Fig. 3a). As the inter-class similarity is not handled in the center loss, overlap can still be observed between clusters as shown in Fig. 3b. In contrast, the proposed island loss deals with the intra-class variations and the inter-class similarities simultaneously. Thus, as depicted in Fig. 3c, the features learned using the island loss form more compact clusters, which are better separated in the feature space, compared with those learned using the softmax loss and the center loss.

### F. A Study of the Distances

To further demonstrate the discriminative power of the proposed island loss, the cosine distances between centers are analyzed for the four databases. For the model using the softmax loss, the cosine distance between the means of each pair of expressions is calculated; while for the model using the center loss or the proposed island loss, the distance between each pair of the learned centers is computed. The final cosine distance is obtained by averaging the distances between all pairs. As shown in Table X, the island loss consistently achieves the largest between-center-distance indicating the enlarged inter-class differences on the four databases.

Furthermore, the cosine distance between each sample and its corresponding center is calculated. The final sample-to-center distance is obtained by averaging the distances of all samples. As shown in Table XI, the proposed island loss consistently produces the smallest sample-to-center distance on the four datasets, which demonstrate the effectiveness of the proposed island loss in reducing the intra-class variation.

## V. CONCLUSION

In this work, a novel island loss is proposed for CNNs to enhance the discriminative power of learned deep features. Specifically, the proposed island loss pulls the samples towards their corresponding class centers to achieve intra-class compactness and at the same time, pushes the centers away from each other to make the clusters as isolated "islands". Experimental results on three posed facial expression datasets and, more importantly, a spontaneous facial expression dataset have demonstrated that the proposed IL-CNN outperforms the baseline CNNs with the traditional softmax loss or the center loss and achieves better or at least comparable performance compared with the state-of-the-art methods for facial expression recognition. As shown in the experiments, the proposed island loss increases the inter-class differences consistently on the four databases as indicated by the increased cosine distance between the class centers. Meanwhile, the intra-class variations are further reduced as compared to the one using the center loss.

Note that the proposed island loss is a general loss function for CNNs and is ready to be adopted by other advanced network structures for various computer vision and machine learning problems.